\documentclass[11pt,a4paper]{article}
\usepackage{nodalida2021}
\usepackage{url}
\usepackage{latexsym}
\usepackage{graphicx}
\usepackage{hyperref}
\usepackage{times}
\usepackage{xcolor}
\usepackage{latexsym}
\usepackage{csquotes}
\usepackage{multicol,array,booktabs,siunitx,multirow,dcolumn}
\usepackage{tikz-dependency,subcaption,float}

\renewrobustcmd{\bfseries}{\fontseries{b}\selectfont}
\usepackage{microtype}
\aclfinalcopy % Uncomment this line for the final submission
\definecolor{deppink}{HTML}{a92b7a}
\definecolor{depgreen}{HTML}{679033}
\definecolor{deporange}{HTML}{be6320}
\definecolor{depblue}{HTML}{165f77}%7572af}
\definecolor{depred}{HTML}{b12019}

\title{What Taggers Fail to Learn, Parsers Need the Most}
\author{Mark Anderson\qquad Carlos G\'{o}mez-Rodr\'{i}guez\\
  Universidade da Coru\~na, CITIC \\
  FASTPARSE Lab, LyS Research Group, \\ 
  Departamento de Ciencias de la Computaci\'{o}n y Tecnolog\'{i}as de la Informaci\'{o}n \\
  \{\texttt{m.anderson,carlos.gomez}\}\texttt{@udc.es}}

\date{}
\definecolor{deppink}{HTML}{a92b7a}
\definecolor{depgreen}{HTML}{679033}
\definecolor{deporange}{HTML}{be6320}
\definecolor{depblue}{HTML}{165f77}%7572af}
\definecolor{depred}{HTML}{b12019}
\definecolor{depgrey}{HTML}{6c8093}
\definecolor{white}{HTML}{ffffff}

\makeatletter

\newcommand*{\@rowstyle}{}

\newcommand*{\rowstyle}[1]{% sets the style of the next row
  \gdef\@rowstyle{#1}%
  \@rowstyle\ignorespaces%
}

\newcolumntype{=}{% resets the row style
  >{\gdef\@rowstyle{}}%
}

\newcolumntype{+}{% adds the current row style to the next column
  >{\@rowstyle}%
}

\makeatother

\begin{document}
\maketitle
\begin{abstract} 
We present an error analysis of neural UPOS taggers to evaluate why using gold standard tags has such a large positive contribution to parsing performance while using predicted UPOS tags either harms performance or offers a negligible improvement. We evaluate what neural dependency parsers implicitly learn about word types and how this relates to the errors taggers make to explain the minimal impact using predicted tags has on parsers. We also present a short analysis on what contexts result in reductions in tagging performance. We then mask UPOS tags based on errors made by taggers to tease away the contribution of UPOS tags which taggers succeed and fail to classify correctly and the impact of tagging errors.
\end{abstract}
\section{Introduction}
Part-of-speech (POS) tags have commonly been used as input features for dependency parsers. They were especially useful for non-neural implementations \cite{voutilainen1998does,dalrymple2006much,alfared2012pos}. However, the efficacy of POS tags for neural network dependency parsers is less apparent especially when utilising character embeddings \cite{ballesteros2015improved,de2017raw}. Universal POS (UPOS) tags have still been seen to improve parsing performance but only if the predicted tags come from a sufficiently accurate tagger \cite{dozat2017stanford}. 

Typically using predicted POS tags has offered a nominal increase in performance or has had no impact at all. \citet{smith2018investigation} undertook a thorough systematic analysis of the interplay of UPOS tags, character embeddings, and pre-trained word embeddings for multi-lingual Universal Dependency (UD) parsing and found that tags offer a marginal improvement for their transition based parser. However, \citet{zhang2020pos} found that the only way to leverage POS tags (both coarse and fine-grained) for English and Chinese dependency parsing was to utilise them as an auxiliary task in a multi-task framework. Further, \citet{anderson-gomez-2020-frailty} investigated the impact UPOS tagging accuracy has on graph-based and transition-based parsers and found that a prohibitively high tagging accuracy was needed to utilise predicted UPOS tags. Here we investigate whether dependency parsers inherently learn similar word type information to taggers,
and therefore can only benefit from the hard to predict tags that taggers fail to capture. We also investigate what makes them hard to predict. 
\section{Methodology}
We performed two experiments. The first was an attempt to compare what biaffine parsers learn about UPOS tags by fine-tuning them with tagging information and comparing their errors with those from normally trained UPOS taggers. The second experiment attempted to evaluate the impact tagging errors have by either masking errors or using the gold standard tags for erroneously predicted tags while masking all other tags. 
\paragraph{Data} We took a subset of UD v2.6 treebanks consisting of 11 languages, all of which are from different language families \cite{ud26}: Arabic PADT (ar), Basque BDT (eu), Finnish TDT (fi), Indonesian GSD (id), Irish IDT (ga), Japanese GSD (ja), Korean Kaist (ko), Tamil TTB (ta), Turkish IMST (tr), Vietnamese VTB (vi), and Wolof WTB (wo). We used pre-trained word embeddings from fastText (for Wolof we had to use the previous Wiki version) \citep{bojanowski2017enriching,grave2018learning}. We compressed the word embeddings to 100 dimensions with PCA.

\paragraph{Experiment 1: Error crossover} We trained parsers and taggers on the subset of UD treebanks described above. We then took the parser network and replaced the biaffine structure with a multi-layer perceptron (MLP) to predict UPOS tags. We froze the network except for the MLP and fine-tuned the MLP with one epoch of learning, which is similar to the process used in \citet{vania-etal-2019-systematic}. We train for only one epoch to balance training the MLP to decode what the system already has encoded without giving it the opportunity to encode more information. We repeated this for the tagger networks (replacing their MLP with a randomly initialised MLP) to validate this fine-tuning procedure. We then compared the tagging errors of both the parsers fine-tuned for tagging and the original taggers. We also undertook an analysis of the errors from the normal taggers which included looking at the impact out-of-vocabulary, POS tag context, and a narrow syntactic context. We define the contexts in Section \ref{sec:results}. 
\paragraph{Experiment 2: Masked tags}
We then used the output from the taggers from Experiment 1 to train different parsers. We trained parsers using all the predicted tags, using only the gold standard tags the taggers failed to predict (for both the standard taggers and parsers fine-tuned for tagging),
using predicted tags from the standard taggers but masking the errors, and training with all gold standard tags. Note that the respective sets of POS tags were used at both training and inference time. We also trained parsers with no tags as a baseline.
\begin{table}[tb!]
\small
    \centering
    \begin{tabular}{lccc}
    \toprule
         & Tagger  & Tagger-FT & Parser \\ \midrule
    \textbf{Arabic} & 96.71 & 96.52 & 93.73\\     
    \textbf{Basque} & 95.35 & 95.18 & 88.09\\     
    \textbf{Finnish} & 96.92 & 96.62 & 92.24\\     
    \textbf{Indonesian} & 93.72 & 93.79 & 91.98\\     
    \textbf{Irish} & 92.84 & 92.80 & 88.24\\     
    \textbf{Japanese} & 97.94 & 97.85 & 92.80\\     
    \textbf{Korean} & 95.09 & 94.26 & 86.93\\     
    \textbf{Tamil} & 89.29 & 87.28 & 75.41\\     
    \textbf{Turkey} & 95.10 & 94.98 & 86.14\\     
    \textbf{Vietnamese} & 87.85 & 87.63 & 83.40\\     
    \textbf{Wolof} & 93.85 & 93.79 & 85.81\\     
    \bottomrule
    \end{tabular}
    \caption{Tagging accuracies for tagger trained normally (Tagger), ``fine-tuning'' a newly initialised MLP for the trained taggers (Tagger-FT), and for parsers fine-tuned to predict tags (Parser).}
    \label{tab:tagging_scores}
\end{table}
\paragraph{Network details} Both the taggers and parsers use pre-trained word embeddings and randomly-initialised character embeddings. The parsers use UPOS tag embeddings as specified in the experimental details. The character and tag embeddings are randomly initialised. The parsers consist of the embedding layer followed by BiLSTM layers and then a biaffine mechanism \cite{dozat20161}. The taggers are similar but with an MLP following the BiLSTMs instead.  We ran a small hyperparameter search using fi, ga, tr, and wo and using their respective development data. This resulted in 3 BiLSTM layers with 200 nodes, 100 dimensions for each embedding type with 100 dimension input to the character LSTM. The arc MLP of the biaffine structure
had 100 dimensions, 50 for the relation MLP.
%was 100 and the relation MLP was 50. 
Dropout was 0.33 for all layers. Learning rate was 2$\times10^{-3}$, $\beta_{1}$ and $\beta_{2}$ were 0.9, batch size was 30, and we trained both taggers and parsers for 200 epochs but with early stopping if no improvement was seen after 20 epochs. Models were selected based on the performance on the development set.
\begin{figure}[tb!]
    \centering
    \includegraphics[width=0.5\linewidth]{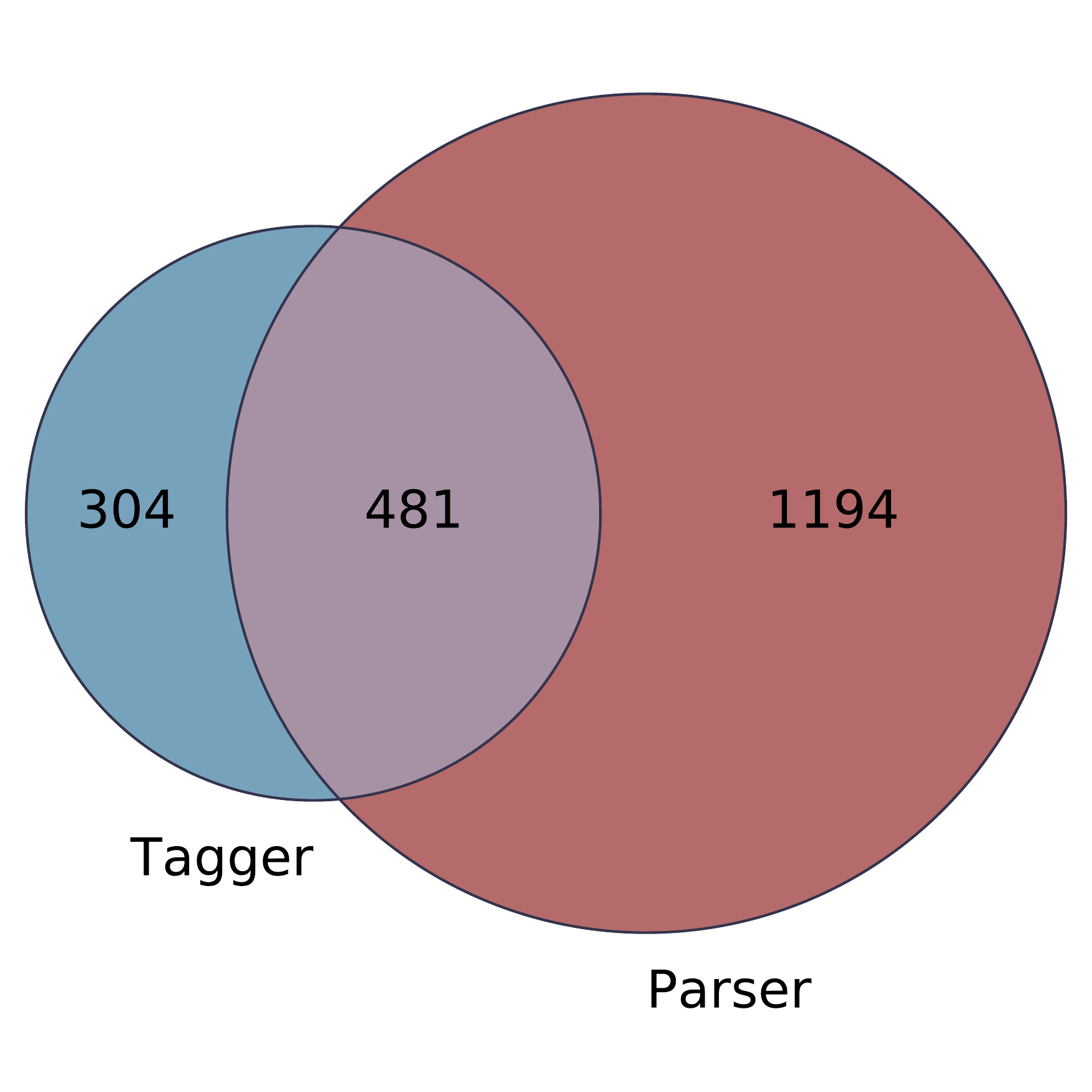}
    \caption{Average union of tagging errors for parser fine-tuned for tagging and fully-trained tagger (standard deviation: 159 for tagger error, 715 for parser, and 242 for union).}
    \label{fig:error_union}
\end{figure}
\begin{table}[b!]
\small
    \centering
    \tabcolsep=0.18cm
    \begin{tabular}{lrrrr}
    \toprule
         & All  &Open & Closed & Other \\ \midrule
    \textbf{Tagger}  &8,637 & 6,434 & 1,867 & 336\\    
    \textbf{Parser} &18,426 & 15,181 & 2,816 & 429\\\midrule
    \textbf{Total} &171,373 & 101,965 & 46,362 & 23,046 \\
    \bottomrule
    \end{tabular}
    \caption{Error (Parser, Tagger) and total (Total) counts across all data per word class of gold tag.}
    \label{tab:class}
\end{table}
\begin{table*}[htpb!]
\small
     \tabcolsep=.19cm
     \centering
    \begin{tabular}{l
    rrrrr
    rr}
    \toprule
    & \multicolumn{5}{c}{Error Types} & Errors & Tokens \\\midrule
\textbf{ar} & noun$\rightarrow$x 197 &  x$\rightarrow$noun 139 &  noun$\rightarrow$adj 108 &  adj$\rightarrow$x 78 &  adj$\rightarrow$noun 60 &  931  &  28.3K \\
\textbf{eu} & propn$\rightarrow$noun 145 &  verb$\rightarrow$aux 113 &  noun$\rightarrow$adj 101 &  aux$\rightarrow$verb 100 &  adj$\rightarrow$noun 94 &  1134  &  24.4K \\
\textbf{fi} & propn$\rightarrow$noun 56 &  noun$\rightarrow$propn 53 &  noun$\rightarrow$adj 43 &  adj$\rightarrow$noun 39 &  noun$\rightarrow$verb 37 &  649  &  21.1K \\
\textbf{id} & propn$\rightarrow$noun 147 &  noun$\rightarrow$propn 92 &  adj$\rightarrow$noun 47 &  noun$\rightarrow$adj 34 &  verb$\rightarrow$noun 23 &  740  &  11.8K \\
\textbf{ga} & propn$\rightarrow$noun 184 &  noun$\rightarrow$propn 53 &  noun$\rightarrow$adj 53 &  adj$\rightarrow$noun 38 &  noun$\rightarrow$pron 36 &  724  &  10.1K \\
\textbf{ja} & noun$\rightarrow$adv 52 &  propn$\rightarrow$noun 24 &  noun$\rightarrow$adj 22 &  adj$\rightarrow$noun 22 &  aux$\rightarrow$verb 20 &  269  &  13.0K \\
\textbf{ko} & noun$\rightarrow$propn 252 &  propn$\rightarrow$noun 145 &  verb$\rightarrow$adj 133 &  aux$\rightarrow$verb 78 &  cconj$\rightarrow$sconj 75 &  1394  &  28.4K \\
\textbf{ta} & noun$\rightarrow$propn 24 &  aux$\rightarrow$verb 22 &  propn$\rightarrow$noun 17 &  noun$\rightarrow$verb 12 &  adj$\rightarrow$adp 12 &  213  &  2.0K \\
\textbf{tr} & noun$\rightarrow$adj 54 &  propn$\rightarrow$noun 52 &  noun$\rightarrow$verb 37 &  noun$\rightarrow$propn 35 &  adv$\rightarrow$adj 31 &  491  &  10.0K \\
\textbf{vi} & noun$\rightarrow$verb 201 &  verb$\rightarrow$noun 152 &  noun$\rightarrow$adj 151 &  verb$\rightarrow$adj 140 &  verb$\rightarrow$x 83 &  1452  &  12.0K \\
\textbf{wo} & noun$\rightarrow$propn 71 &  verb$\rightarrow$noun 57 &  pron$\rightarrow$det 46 &  noun$\rightarrow$verb 38 &  verb$\rightarrow$aux 30 &  640  &  10.4K \\
    \bottomrule
    \end{tabular}
    \caption{Top 5 most common errors and their number of occurrences for each treebank. Also shown are the total number of errors and token count for each treebank.}
    \label{tab:prominent_errors}
\end{table*}
\section{Results and discussion}\label{sec:results}
\paragraph{Experiment 1: Error crossover} Table \ref{tab:tagging_scores} shows the tagging performance for the normally trained taggers, the re-fine-tuned taggers, and the fine-tuned parser taggers. The re-fine-tuned taggers achieve relatively similar performance to the original taggers, which suggests that this procedure does allow us to develop a decoder that captures what the BiLSTM and embedding layers learn about UPOS tags without adding new information. Clearly more training would likely improve the parsers fine-tuned for tagging, but it would be less clear if that would be extracting information the parser previously learnt or adding more information via MLP weights.

\begin{table}[b!]
    \centering
    \small
    \tabcolsep=.3cm
    \begin{tabular}{lcrrr}
    \toprule
    & \multicolumn{2}{c}{F1-score} \\
    & Tagger & \multicolumn{1}{c}{Parser} & \multicolumn{1}{c}{Tokens} & Class\\
    \midrule
    PUNCT& 99.94& 99.93 & 19.9K & \multirow{3}{*}{Other}\\
    SYM& 97.83& 0.00 & 0.2K\\
    X& 76.37& 54.51 & 2.6K \\
    \midrule
    ADJ& 87.98& 74.98 & 9.4K & \multirow{6}{*}{Open}\\
    ADV& 93.94& 89.97 & 8.5K \\
    INTJ& 40.91& 0.00 & 0.1K \\
    NOUN& 95.49& 94.63 & 43.7K \\
    PROPN& 90.21& 57.49 & 9.0K \\
    VERB& 94.80& 94.05 & 21.5 \\
    \midrule
    ADP& 97.77& 94.14 & 9.9K & \multirow{8}{*}{Closed}\\
    AUX& 96.37& 93.65 & 6.8K \\
    CCONJ& 96.30& 94.29 & 7.3K \\
    DET& 94.73& 86.88 & 4.2K \\
    NUM& 93.96& 78.12 & 4.4K \\
    PART& 90.49& 76.88 & 1.7K \\
    PRON& 96.31& 72.46 & 6.0K \\
    SCONJ& 93.15& 91.22 & 3.2K \\
        \bottomrule
    \end{tabular}
    \caption{F1-score for separate tags clustered by word type class with "Other" at the top, "Open" in the middle, and "Closed" at the bottom for all tokens in the collection of treebanks used. Also reported are the total number of tokens for each tag type present across all treebanks (Tokens).}
    \label{tab:tag-f1}
\end{table}

Figure \ref{fig:error_union} shows the average cross-over of specific error occurrences for the two systems, where only 38\% of the tagger's errors don't occur for the parser. Table \ref{tab:class} shows the breakdown of errors from each system by word type class for all treebanks. The ratio of the errors is substantially different for each class: 0.42 for \textit{open}, 0.66 for \textit{closed}, 0.78 for \textit{other}. This perhaps suggests that the parser has a tendency to learn more syntactically fixed word types than open types. Table \ref{tab:tag-f1} shows the F1-score for each UPOS for both systems. 
For the most part the parser is pretty close to the tagger for open class tags, except for \texttt{INTJ} which the parser never predicts, \texttt{PROPN} (32.7 less for the parser), and to a lesser extent \texttt{ADJ} (13.0 less). 
Table \ref{tab:prominent_errors} shows the top 5 most common errors per treebank for the normal taggers where \texttt{PROPN} appears in 15 error types and \texttt{ADJ} appears in 19 out of 55. This prevalence combined with the parsers' poor performance for these tags suggests that errors containing these tags are especially impactful for parsers when using predicted UPOS. However, it could also be that the parsers perform poorly on predicting \texttt{PROPN} tags as they occur in similar syntactic roles as \texttt{NOUN} tokens and as such aren't as important for syntactic analysis. 

For the closed class type tags, again the parser performs similarly to the tagger but obtains a few points less except for \texttt{DET}, \texttt{NUM}, \texttt{PART}, and \texttt{PRON} with drops for parser scores of 7.9, 15.8, 13.6, and 23.9, respectively. However, of these 4 tags, only \texttt{PRON} and \texttt{DET} appear in the most common errors and only twice and once, respectively. The most common tag to appear in an error is \texttt{NOUN} occurring 41 times, but there is less than one point in difference between the tagger's performance and the parser's for \texttt{NOUN}. Of these 41 appearances, 14 co-occur with \texttt{ADJ} and 15 with \texttt{PROPN} with a fairly even split of mis-tagging \texttt{NOUN} as either of these tags or the other way around. So generally \texttt{NOUN} tokens are fairly easy to tag, but the times where the tagger fails are typically where there is confusion with \texttt{ADJ} and \texttt{PROPN} tags.
\begin{figure*}[thpb!]
    \centering
    \includegraphics[width=0.825\linewidth]{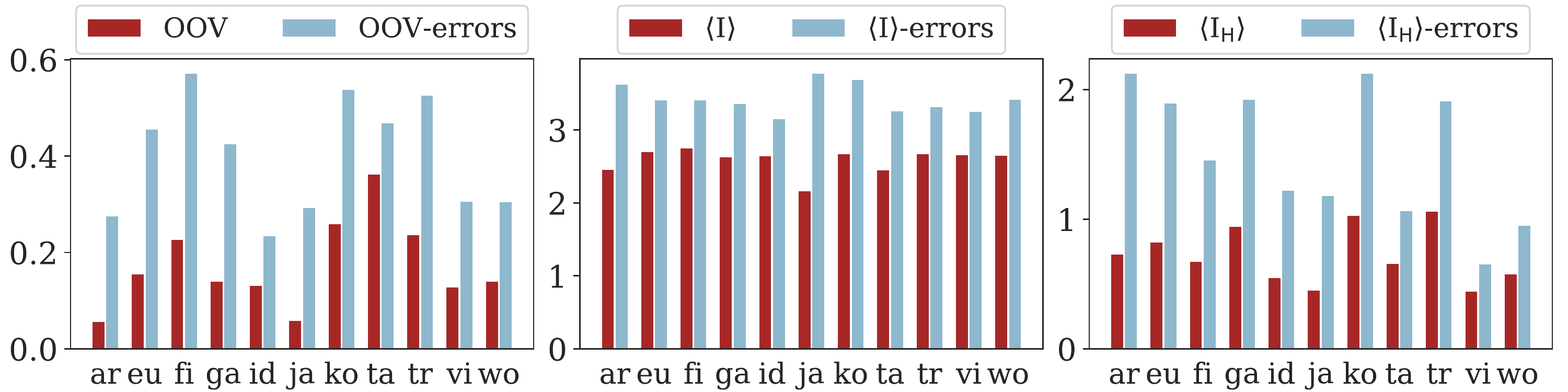}
    \caption{Measurements of all tags (red) and error (blue) tags for OOV proportion, POS bigram surprisal ($\langle$I$\rangle$, $\langle$I$\rangle$-errors),  and head POS and relation surprisal ($\langle$I$_{\textrm{H}}\rangle$, $\langle$I$_{\textrm{H}}\rangle$-errors).}
    \label{fig:stats}
\end{figure*}
Figure \ref{fig:stats} shows statistical metrics of the taggers' errors. First we show the proportion of out-of-vocabulary (OOV) word forms for all tokens and also the tokens where the tagger makes an error. Consistently across all treebanks the OOV proportion is considerably higher for tokens erroneously tagged. Second we report the mean UPOS surprisal. For a given UPOS tag, $\theta_{n}$ for token n, the surprisal of that UPOS tag in a given context, $c_{k}$ is given as:
\begin{equation}\label{eq:surprisal}
    I(\theta_{n}) = -\log_{2}p(\theta_{n}|c_{k})
\end{equation}
where we use a bigram context:
\begin{equation}
c_{k} = (\theta_{n-2},\theta_{n-1})
\end{equation}
Then the mean surprisal, $\langle I\rangle$, over a sample of tokens is given as:
\begin{equation}\label{eq:mean_surprisal}
\langle I\rangle = \frac{1}{N}\sum_{n\in N}I(\theta_{n})
\end{equation}
where $N$ is the number of tokens in the sample. Again, the mean tag surprisal is substantially different across all treebanks for the tokens where the tagger makes a mistake in comparison to the average over the entire treebank. Finally we report the mean surprisal of UPOS but with the context of its head's tag and the syntactic relation joining the two tokens, such that $c_{k}$ is defined as:
\begin{equation}
c_{k} = (\theta_{head},rel)
\end{equation}
%Otherwise, Equations \ref{eq:surprisal} and \ref{eq:mean_surprisal} are the same. 
The difference between the error sub-sample and the whole treebank is starker for the head-relation surprisal, suggesting that the tagger struggles more when the syntactic structure is uncommon.
\begin{table}[b!]
\small
    \centering
      \tabcolsep=.18cm
    \begin{tabular}{lcccccc}
    \toprule
         & None & Pred. & M$\neg$E$_\textrm{T}$ &  M$\neg$E$_\textrm{P}$ & M$\forall$E$_\textrm{T}$ &Gold \\ \midrule
\textbf{ar}  & 83.29  & 82.87 & 84.17 & 84.06 & 84.45 & 84.73 \\
\textbf{eu}  & 81.12  & 81.14 & 82.33 & 82.62 & 83.13 & 84.45 \\
\textbf{fi}  & 85.96  & 86.04 & 86.88 & 87.09 & 87.61 & 88.80 \\
\textbf{id}  & 79.04  & 78.95 & 82.20 & 82.69 & 81.08 & 82.95 \\
\textbf{ga}  & 76.13  & 76.57 & 76.62 & 76.65 & 77.46 & 77.90 \\
\textbf{ja}  & 93.15  & 92.72 & 94.41 & 94.38 & 94.39 & 95.30 \\
\textbf{ko}  & 85.40  & 85.86 & 87.53 & 87.82 & 87.44 & 88.52 \\
\textbf{ta}  & 65.61  & 64.50 & 70.24 & 66.67 & 66.01 & 71.95 \\
\textbf{tr}  & 66.67  & 67.68 & 67.62 & 67.66 & 67.84 & 68.86 \\
\textbf{vi}  & 58.43  & 60.09 & 65.42 & 66.75 & 65.18 & 70.87 \\
\textbf{wo}  & 77.87  & 78.49 & 82.03 & 81.39 & 81.11 & 85.41 \\\midrule
\textbf{avg} & 77.52 & 77.72 & 79.95 & 79.80 & 79.61  & 81.79 \\
    \bottomrule
    \end{tabular}
    \caption{LAS parser performance with no tags (None), with predicted tags (Pred), gold standard tags but with all tags masked except those the respective taggers predicted wrong (M$\neg$E$_\textrm{T}$), similarly for the tagging errors from the fine-tuned parser (M$\neg$E$_\textrm{P}$), masking the errors from the tagger (M$\forall$E$_\textrm{T}$), and finally using all gold standard tags.}
    \label{tab:las}
\end{table}
\paragraph{Experiment 2: Masked tags} Table \ref{tab:las} shows the labelled attachment scores for parsers with varying types of UPOS input. First we use the predicted output from the normal taggers from Experiment 1 (Pred) and unlike \citet{anderson-gomez-2020-frailty} we observe a slight increase over using no UPOS tags. However, using predicted tags isn't universally beneficial. Arabic, Indonesian, Japanese, and Tamil all perform better with no tags. 

We then used gold standard tags but masking the tags that the taggers correctly predicted to test if the erroneous tags are particularly useful. We did this for the normal taggers (M$\neg$E$_\textrm{T}$) and also for the fine-tuned parsers (M$\neg$E$_\textrm{P}$). The average increase for both is about 2.5 over the no tag baseline and over 2 points better than using predicted tags. Also, the improvement is universal with at least a small increase in performance over using predicted UPOS tags. Interestingly the smaller set from the tagger outperforms the larger set from the parser by 0.15, suggesting that what both the taggers and the parsers fail to capture is more important than the errors unique to the parsers. We then masked the errors from the taggers (M$\forall$E$_\textrm{T}$) to test if avoiding adding errors would still be beneficial. The performance is almost 2 points better than using the predicted tags and again an increase is observed for all treebanks. This could be of use, as it is easy to envisage a tagger which learns to predict tags when a prediction is clear and to predict nothing when the probability is low. Finally, using gold standard tags is nearly 2 points better on average than the best masked tag model, which suggests that to fully utilise the information in the final few percentage that taggers miss, the full set of easy to predict tags are needed.
\section{Conclusion}
We have presented results which suggest that parsers do learn something of word types and that what taggers fail to learn is needed to augment that knowledge. We have evaluated the nature of typical tagging errors for a diverse subset of UD treebanks and highlighted consistent error types and also what statistical features they have compared to the average measurement across all tokens in a treebank. We have shown that it would be more beneficial to implement taggers to not only predict tags but also decide when to do so, as the errors undermine anything gained from using predicted tags for dependency parsers.  Note that while we only used one parser system, the original paper \cite{anderson-gomez-2020-frailty} which prompted this work observed similar behaviour with regard to predicted UPOS tags for both the system used here (graph-based) and a neural transition-based parser, suggesting that the results discussed here might extend to other parsing systems. And while it is true that we have only investigated one POS tagger system, we feel we have been careful in not making egregiously grand claims of the universality of our findings: it is merely one data point to be considered amongst many.
\section*{Acknowledgments}
This work has received funding from the European Research Council (ERC), under the European Union's Horizon 2020 research and innovation programme (FASTPARSE, grant agreement No 714150), from MINECO (ANSWER-ASAP, TIN2017-85160-C2-1-R), from Xunta de Galicia (ED431C 2020/11), and from Centro de Investigación de Galicia ``CITIC'', funded by Xunta de Galicia and the European Union (ERDF - Galicia 2014-2020 Program), by grant ED431G 2019/01. The authors would also like to the thank the reviewers for their suggestions and criticisms. 
%\clearpage
\bibliography{eacl2021}
\bibliographystyle{acl_natbib}
%\clearpage
%\appendix
%\onecolumn
%\section{Appendix}

\end{document}